\documentclass{article}

\makeatletter

\usepackage[verbose=true,letterpaper]{geometry}
\AtBeginDocument{
  \newgeometry{
    textheight=9in,
    textwidth=6.5in,
    top=1in,
    headheight=14pt,
    headsep=25pt,
    footskip=30pt
  }
}

\newcommand{\headeright}{A Preprint}
\newcommand{\undertitle}{A Preprint}
\newcommand{\shorttitle}{\@title}

\usepackage{fancyhdr}
\fancyheadoffset{0pt}
\def\keywordname{{\bfseries \emph{Keywords}}}%
\def\keywords#1{\par\addvspace\medskipamount{\rightskip=0pt plus1cm
\def\and{\ifhmode\unskip\nobreak\fi\ $\cdot$
}\noindent\keywordname\enspace\ignorespaces#1\par}}

\renewcommand{\normalsize}{%
  \@setfontsize\normalsize\@xpt\@xipt
  \abovedisplayskip      7\p@ \@plus 2\p@ \@minus 5\p@
  \abovedisplayshortskip \z@ \@plus 3\p@
  \belowdisplayskip      \abovedisplayskip
  \belowdisplayshortskip 4\p@ \@plus 3\p@ \@minus 3\p@
}
\normalsize
\renewcommand{\small}{%
  \@setfontsize\small\@ixpt\@xpt
  \abovedisplayskip      6\p@ \@plus 1.5\p@ \@minus 4\p@
  \abovedisplayshortskip \z@  \@plus 2\p@
  \belowdisplayskip      \abovedisplayskip
  \belowdisplayshortskip 3\p@ \@plus 2\p@   \@minus 2\p@
}
\renewcommand{\footnotesize}{\@setfontsize\footnotesize\@ixpt\@xpt}
\renewcommand{\scriptsize}{\@setfontsize\scriptsize\@viipt\@viiipt}
\renewcommand{\tiny}{\@setfontsize\tiny\@vipt\@viipt}
\renewcommand{\large}{\@setfontsize\large\@xiipt{14}}
\renewcommand{\Large}{\@setfontsize\Large\@xivpt{16}}
\renewcommand{\LARGE}{\@setfontsize\LARGE\@xviipt{20}}
\renewcommand{\huge}{\@setfontsize\huge\@xxpt{23}}
\renewcommand{\Huge}{\@setfontsize\Huge\@xxvpt{28}}

\providecommand{\section}{}
\renewcommand{\section}{%
  \@startsection{section}{1}{\z@}%
                {-2.0ex \@plus -0.5ex \@minus -0.2ex}%
                { 1.5ex \@plus  0.3ex \@minus  0.2ex}%
                {\large\bf\raggedright}%
}
\providecommand{\subsection}{}
\renewcommand{\subsection}{%
  \@startsection{subsection}{2}{\z@}%
                {-1.8ex \@plus -0.5ex \@minus -0.2ex}%
                { 0.8ex \@plus  0.2ex}%
                {\normalsize\bf\raggedright}%
}
\providecommand{\subsubsection}{}
\renewcommand{\subsubsection}{%
  \@startsection{subsubsection}{3}{\z@}%
                {-1.5ex \@plus -0.5ex \@minus -0.2ex}%
                { 0.5ex \@plus  0.2ex}%
                {\normalsize\bf\raggedright}%
}
\providecommand{\paragraph}{}
\renewcommand{\paragraph}{%
  \@startsection{paragraph}{4}{\z@}%
                {1.5ex \@plus 0.5ex \@minus 0.2ex}%
                {-1em}%
                {\normalsize\bf}%
}
\providecommand{\subparagraph}{}
\renewcommand{\subparagraph}{%
  \@startsection{subparagraph}{5}{\z@}%
                {1.5ex \@plus 0.5ex \@minus 0.2ex}%
                {-1em}%
                {\normalsize\bf}%
}

\newlength{\@abovecaptionskip}
\newlength{\@belowcaptionskip}

\renewenvironment{table}
  {\setlength{\abovecaptionskip}{\@belowcaptionskip}%
   \setlength{\belowcaptionskip}{\@abovecaptionskip}%
   \@float{table}}
  {\end@float}

\renewcommand{\footnoterule}{\kern-3\p@ \hrule width 12pc \kern 2.6\p@}
\def\@listi  {\leftmargin\leftmargini}
\def\@listii {\leftmargin\leftmarginii
              \labelwidth\leftmarginii
              \advance\labelwidth-\labelsep
              \topsep  2\p@ \@plus 1\p@    \@minus 0.5\p@
              \parsep  1\p@ \@plus 0.5\p@ \@minus 0.5\p@
              \itemsep \parsep}
\def\@listiii{\leftmargin\leftmarginiii
              \labelwidth\leftmarginiii
              \advance\labelwidth-\labelsep
              \topsep    1\p@ \@plus 0.5\p@ \@minus 0.5\p@
              \parsep    \z@
              \partopsep 0.5\p@ \@plus 0\p@ \@minus 0.5\p@
              \itemsep \topsep}
\def\@listiv {\leftmargin\leftmarginiv
              \labelwidth\leftmarginiv
              \advance\labelwidth-\labelsep}
\def\@listv  {\leftmargin\leftmarginv
              \labelwidth\leftmarginv
              \advance\labelwidth-\labelsep}
\def\@listvi {\leftmargin\leftmarginvi
              \labelwidth\leftmarginvi
              \advance\labelwidth-\labelsep}

\providecommand{\maketitle}{}
\renewcommand{\maketitle}{%
  \par
  \begingroup
    \renewcommand{\thefootnote}{\fnsymbol{footnote}}
    \long\def\@makefntext##1{%
      \parindent 1em\noindent
      \hbox to 1.8em{\hss $\m@th ^{\@thefnmark}$}##1
    }
    \thispagestyle{empty}
    \@maketitle
    \@thanks
  \endgroup
  \let\maketitle\relax
  \let\thanks\relax
}

\newcommand{\@toptitlebar}{
  \hrule height 2\p@
  \vskip 0.25in
  \vskip -\parskip%
}
\newcommand{\@bottomtitlebar}{
  \vskip 0.29in
  \vskip -\parskip
  \hrule height 2\p@
  \vskip 0.09in%
}

\providecommand{\@maketitle}{}
\renewcommand{\@maketitle}{%
  \vbox{%
    \hsize\textwidth
    \linewidth\hsize
    \vskip 0.1in
    \@toptitlebar
    \centering
    {\LARGE\sc \@title\par}
    \@bottomtitlebar
    \textsc{\undertitle}\\
    \vskip 0.1in
    \def\And{%
      \end{tabular}\hfil\linebreak[0]\hfil%
      \begin{tabular}[t]{c}\bf\rule{\z@}{24\p@}\ignorespaces%
    }
    \def\AND{%
      \end{tabular}\hfil\linebreak[4]\hfil%
      \begin{tabular}[t]{c}\bf\rule{\z@}{24\p@}\ignorespaces%
    }
    \begin{tabular}[t]{c}\bf\rule{\z@}{24\p@}\@author\end{tabular}%
  \vskip 0.4in \@minus 0.1in \center{\@date}   \vskip 0.2in
  }
}

\newcommand{\ftype@noticebox}{8}
\newcommand{\@notice}{%
  \enlargethispage{2\baselineskip}%
  \@float{noticebox}[b]%
    \footnotesize\@noticestring%
  \end@float%
}

\renewenvironment{abstract}
{
  \centerline
  {\large \bfseries \scshape Abstract}
  \begin{quote}
}
{
  \end{quote}
}
\makeatother

\usepackage{graphicx}
\usepackage{booktabs}
\usepackage{amsmath}
\usepackage{amssymb}
\usepackage{xcolor}
\usepackage{float}
\usepackage[numbers,sort&compress]{natbib}
\usepackage[hidelinks]{hyperref}
\usepackage{microtype}
\usepackage{placeins}

\graphicspath{{figures/}}

\newcommand{\method}{\textsc{SR-JEPA}}
\newcommand{\pred}{g_{\phi}}
\newcommand{\enc}{f_{\theta}}
\newcommand{\tenc}{f_{\bar\theta}}

\title{\textbf{SR-JEPA: Learning Predictive Latent State in 3D Scenes}}
\author{%
  Zihan Zhou\textsuperscript{1} \qquad Qifu Wen\textsuperscript{2} \qquad Xi Zeng\textsuperscript{1} \\[0.55em]
  \textsuperscript{1}Boston University, Boston, MA, USA \\
  \textsuperscript{2}Shanghai Jiao Tong University, Shanghai, China
}
\date{}

\begin{document}
\maketitle

\begin{abstract}
Joint-embedding predictive architectures learn by predicting latent representations of missing
observations, yet many masked JEPAs are evaluated primarily through the encoders they produce. We ask
what a trained predictive pathway itself infers when an entire entity is absent from a native 3D
scene. We introduce \method, a point-native JEPA for scene-scale point clouds whose original frozen
predictive pathway can be queried at a supplied location. At evaluation, every point of one object is
removed before encoding and replaced by the same shape-free 32-point query at its centroid. Training
uses only self-contained 3D EMA targets---no reconstruction, semantic labels, language, or lifted 2D
features. On 5,953 held-out ARKitScenes objects, the imputed latent reaches 43.13\% semantic-identity
macro accuracy, 22.18 points above the strongest floor. Randomizing the prediction path removes 9.78
points, while substituting matched donor context removes 21.98 points. On 8,570 Sr3D support pairs,
the full latent reaches 41.15 AP; identity decoded from the missing-object latent, combined with
anchor identity and geometry, reaches 39.37 AP, leaving an unresolved 1.78-point residual. These
results reveal a queryable, compositional 3D predictive state: the model completes context-dependent
entity content, which downstream computation combines with metric geometry.
\end{abstract}

\section{Introduction}

Joint-embedding predictive architectures (JEPAs) learn by predicting latent targets compatible with
visible context rather than reconstructing every sensory detail~\citep{lecun2022path,ijepa}.
Foundational masked JEPAs established this principle chiefly through the representations transferred
from their encoders~\citep{ijepa,vjepa}. More recent systems increasingly retain or adapt predictive
modules for transformations, dynamics, reasoning, and control~\citep{iwm,vjepa2,cjepa}. This progress
makes the predictor a scientific object in its own right. What does the original frozen predictive
pathway already infer when its target is not merely masked in patches, but entirely absent?

Native 3D scenes provide a particularly clean setting for this question. Metric coordinates can say
\emph{where} to query while all target evidence that says \emph{what} occupies that location is
removed. A useful predictive state should then recover entity content compatible with the visible
scene, and that content should combine with geometry to support structural decisions. This is the
spatial-completion role envisioned for JEPA-style world models, expressed as a concrete and directly
testable 3D inference problem.

We introduce \method, a point-native JEPA for scene-scale indoor point clouds
(Figure~\ref{fig:arch}). A Point Transformer V3~\citep{ptv3} context encoder processes visible
occupied regions, while a narrow sparse predictor estimates EMA target-encoder representations at
queried locations. Both sides of the objective are learned from native 3D points. No semantic label,
language embedding, input reconstruction, or lifted image-foundation feature enters pretraining.

\begin{figure}[htbp]
  \centering
  \includegraphics[width=\linewidth]{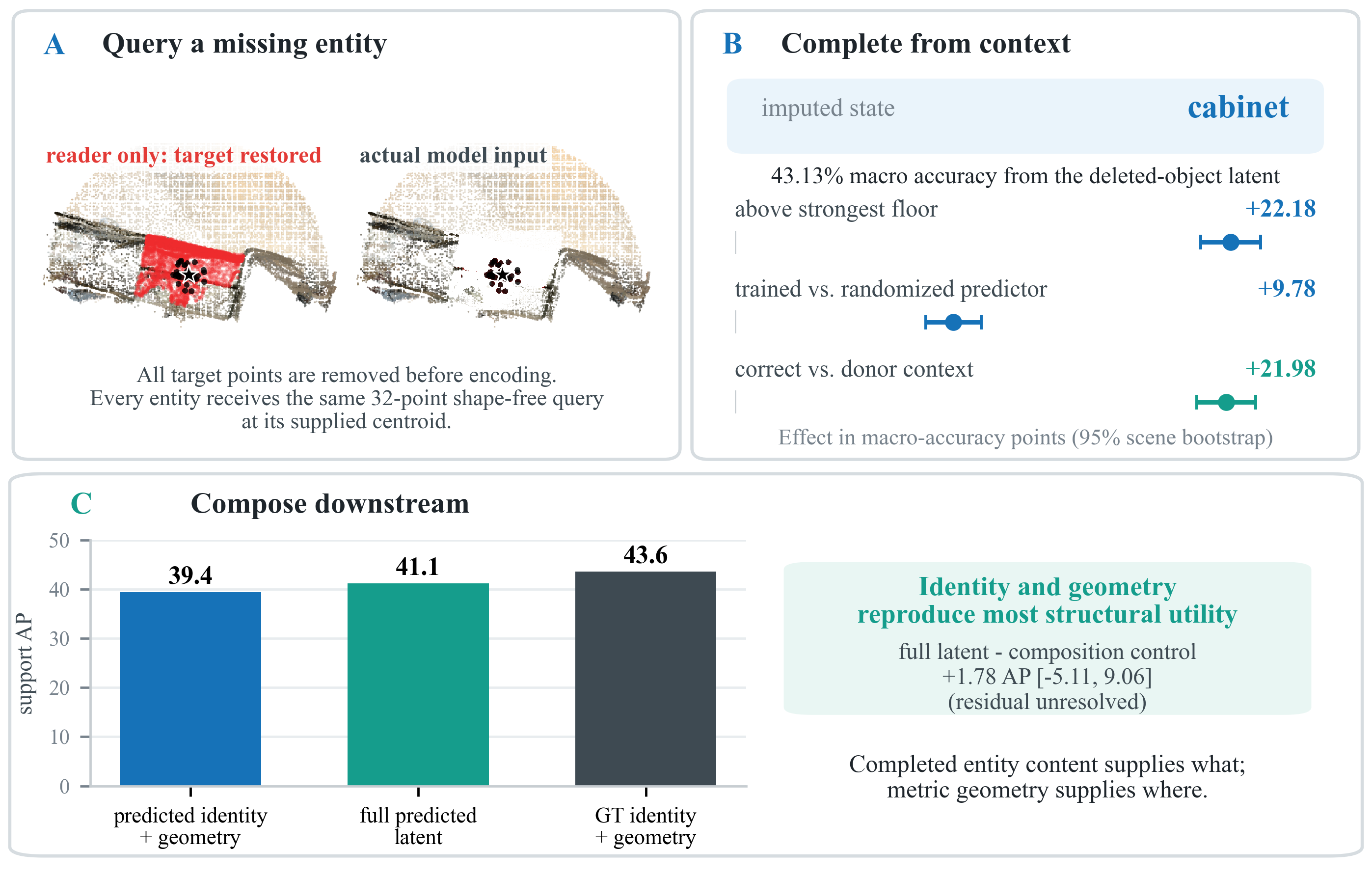}
  \caption{\textbf{Query, completion, and composition expose the content of predictive 3D state.}
  \textbf{(A)} The target is restored in red only for the reader; the model receives the target-deleted
  scene and an identical shape-free query at the supplied centroid. \textbf{(B)} The frozen pathway
  completes entity content that depends materially on predictor training and on the correct scene.
  The cabinet label is illustrative; effects and intervals use all 5,953 endpoint objects.
  \textbf{(C)} Completed target identity, anchor identity, and coordinates reproduce most support
  utility of the full latent; the remaining 1.78-point residual is unresolved. Ground-truth identity
  is a task-susceptibility control, not an input available to \method{} at inference.}
  \label{fig:overview}
\end{figure}

We make the resulting predictive pathway directly testable (Figure~\ref{fig:overview}). At evaluation,
we delete every point of one object before voxelization and insert the same fixed 32-point query at
its supplied centroid. The query contains no target class, dimensions, orientation, silhouette, or
samples. Geometry and random-model floors measure shortcuts; randomized-predictor grafts test the
contribution of the learned prediction path; matched donor scenes test dependence on the correct
context under an unchanged readout. A final composition test asks whether structural utility requires
information beyond completed entity identity and metric geometry.

On 5,953 official ARKitScenes Validation objects from scenes outside pretraining, the imputed state
reaches 43.13\% semantic-identity macro accuracy, 22.18 points above the strongest floor. Randomizing
the prediction path removes 9.78 points, and substituting matched donor context removes 21.98 points.
On 8,570 Sr3D support pairs, the full latent reaches 41.15 AP. A model-level control that combines
identity decoded from the missing-object latent with anchor identity and geometry reaches 39.37 AP;
its 1.78-point residual has a confidence interval crossing zero. The resulting division of labor is
simple and useful: the predictive pathway estimates \emph{what} belongs at the queried location, and
a downstream computation combines entity state with \emph{where} to determine support.

Our contributions are:
\begin{enumerate}
  \item \textbf{A point-native predictive state model for indoor 3D scenes.} \method{} learns
  self-contained 3D EMA targets with occupied-region masking, multiscale latent supervision, a
  sparse scene encoder, and a narrow predictor---without input reconstruction or 2D foundation
  supervision.
  \item \textbf{A frozen predictive interface for missing entity state.} A deletion-first,
  fixed-query protocol removes object shape from the query and tests the original pretrained pathway
  through positional, random-model, predictor, and context contrasts.
  \item \textbf{A controlled account of state content.} External semantic evaluation establishes
  context-dependent entity completion across two training recipes. Joint identity--geometry and
  model-level controls reveal a compositional division between completed entity content, metric
  geometry, and the downstream structural decision.
\end{enumerate}

The present evidence concerns deterministic state completion at a supplied location in a static
scene. Object discovery, temporal transition, action conditioning, rollout, and planning remain
separate capabilities.

\section{Related Work}

\paragraph{Predictor lifecycles in JEPA.}
JEPA predicts representations of compatible inputs rather than reconstructing observations
\citep{lecun2022path}. I-JEPA and V-JEPA use masked latent prediction to learn semantic image and
video representations, with their principal quantitative evidence centered on encoder transfer
\citep{ijepa,vjepa}. Predictor behavior is not absent from this literature: I-JEPA qualitatively
decodes predicted targets, Image World Models broaden prediction to global transformations and
fine-tune the learned model for downstream tasks~\citep{iwm}, and V-JEPA~2 trains a separate causal
predictor over an action-free representation for planning~\citep{vjepa2}. V-JEPA~2.1 develops dense
predictive supervision primarily to improve spatially grounded representations~\citep{vjepa21}.
C-JEPA moves prediction to object-centric histories and futures and evaluates the resulting model for
reasoning and control~\citep{cjepa}. We therefore make no general priority claim for predictor
evaluation. Our complementary question concerns the original frozen pathway in a static native-3D
scene: what state does it produce when a complete entity is absent?

\paragraph{Latent prediction in 3D.}
Point-JEPA applies self-contained EMA targets to object point clouds~\citep{pointjepa}, while
Point2Vec and Hu et al. predict teacher representations or tokens at the same object scale
\citep{point2vec,data2vec,hu3djepa}. At scene scale, MSP performs masked shape and deep-feature
prediction from point-native input and explicitly limits coordinate leakage~\citep{msp}. Locate~3D
pairs a PT-v3 scene encoder with lifted 2D foundation features~\citep{locate3d}; AD-L-JEPA predicts
native LiDAR BEV latents and visualizes masked-region occupancy~\citep{adljepa}; AD-LiST-JEPA adds
future occupancy~\citep{adlistjepa}; and POMA-3D learns point maps with 2D priors~\citep{poma3d}.
\method{} combines point-native indoor scenes, self-contained 3D EMA targets, and a retained
predictor evaluated after complete-object deletion. Its contribution is this testable completion
interface and the resulting account of state content (Appendix
Table~\ref{tab:landscape}).

\paragraph{Completion and controlled state readout.}
Point-MAE and Point-M2AE reconstruct masked point-cloud content~\citep{pointmae,pointm2ae}, while
contrastive and self-distillation systems learn from views, scenes, or cross-modal alignment
\citep{pointcontrast,csc,msc,sonata,concerto}. Dense semantic scene completion also infers unobserved
structure, but trains a task-specific decoder to output occupancy and labels
\citep{sscbench,iso}. Our output instead remains in the pretrained latent space. The supervised head
does not learn occupancy or semantic completion; it is an instrument for identifying content already
present in the pretrained predictive state. This separates our question from task-specific completion
accuracy while retaining complete absence as the evaluation condition.

\paragraph{Relations and controlled probing.}
Sr3D supplies templated object relations for referring-expression grounding, while 3DSSG provides
supervised scene-graph predicates~\citep{referit3d,ssg3d}. Such labels can be strongly predicted from
boxes or class pairs alone~\citep{rel3d,spatialsenseplus}. Following the selectivity principle of
\citet{hewittliang}, we include shuffled labels and architecture-matched random models, but a
predictive system also requires controls that preserve its trained encoder. We therefore randomize
the prediction path and intervene on scene context under the same selected head. For support, we
additionally test identity and geometry jointly and replace oracle target identity with a cross-fitted
posterior decoded from the imputed latent. Together these contrasts distinguish decodability, learned
completion, contextual dependence, and downstream composition.

\section{SR-JEPA: Native 3D Predictive State}

\begin{figure}[htbp]
  \centering
  \includegraphics[width=\linewidth]{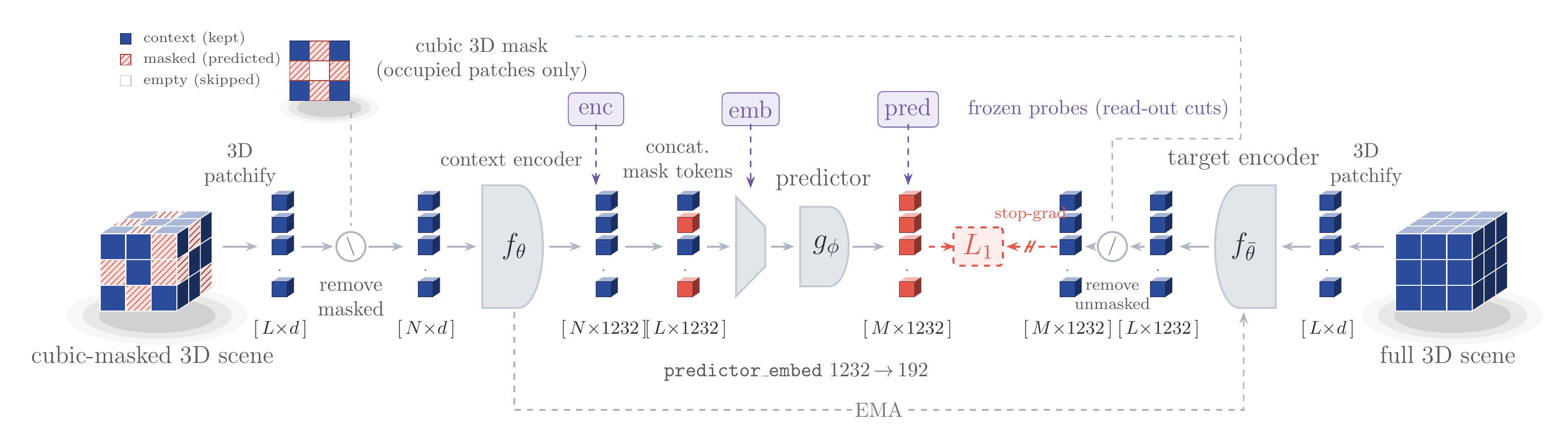}
  \caption{\textbf{SR-JEPA learns point-native 3D state by latent prediction.} A context encoder sees
  visible occupied regions, positional mask tokens query a narrow predictor for missing latents, and
  an EMA target encoder processes the full scene. The marked cuts separate visible encoder state
  (enc), the 192-dimensional predictor embedding (emb), and completed state (pred). We retain the
  frozen pathway and evaluate pred directly after complete-object deletion.}
  \label{fig:arch}
\end{figure}

\subsection{Learning predictive state}

Let a scene be a set of 3D observations
$S=\{(\mathbf{x}_i,\mathbf{c}_i,\mathbf{n}_i)\}_{i=1}^{N}$ containing metric coordinates, color,
and surface normals. A sampled spatial mask $M$ divides the occupied scene into visible observations
$S_{\bar M}$ and target locations $q_M$. The context encoder produces a distributed visible state
$Z_c=\enc(S_{\bar M})$. Conditioned on this state and the query locations, the predictor estimates
\begin{equation}
  \widehat Z_M = \pred(Z_c,q_M).
  \label{eq:predict}
\end{equation}

The regression target $Z_M=\tenc(S)|_M$ is extracted at the corresponding locations from a target
encoder that observes the full scene. Its parameters are an EMA of the context encoder and receive no
gradient. For encoder stage $\ell$, the masked prediction objective is
\begin{equation}
  \mathcal{L}_{\mathrm{mask}}
  = \frac{1}{5}\sum_{\ell=1}^{5}\frac{1}{|M|d_\ell}
    \sum_{i\in M}\left\|\widehat{\mathbf z}^{\ell}_i
      - \operatorname{sg}(\mathbf z^{\ell}_i)\right\|_1.
  \label{eq:loss}
\end{equation}

Here $d_\ell$ is the channel width of stage $\ell$, so every stage contributes equally. We
additionally predict visible context tokens with weight $\lambda=0.1$, following dense predictive
self-supervision in V-JEPA~2.1~\citep{vjepa21}. All targets are produced by the 3D EMA encoder itself:
no object label, language embedding, rendered image feature, or external foundation representation
enters pretraining.

We use \emph{latent scene state} for the set of present-time point features in this learned space.
This is an operational term, not a claim that the representation is a Markov-sufficient state for
control. The current model estimates missing state within one static scene; it does not learn a
temporal transition.

\subsection{Sparse scene encoder and retained predictor}

The context and target encoders are Point Transformer V3 (PT-v3m2) networks~\citep{ptv3} operating
on a $2\,$cm grid. Their five stages have depths $(3,3,3,12,3)$ and widths
$(48,96,192,384,512)$. Features from all stages are upsampled and concatenated to a 1232-dimensional
per-point state. The student receives nine point-native channels: centered metric coordinates, RGB,
and surface normals. The target encoder shares the architecture, is initialized from the student,
uses no stochastic depth, and is updated by EMA with momentum increasing linearly from 0.994 to 1.

The predictor is deliberately smaller: a six-block sparse PT-v3 of width 192 with six attention
heads. A learned linear map projects the 1232-dimensional visible state to the predictor width;
learned mask tokens at $q_M$ request the missing state; and an output projection returns predictions
to the encoder target space. This produces a reusable, queryable prediction interface rather than an
input-space decoder.

The three labeled readout locations in Figure~\ref{fig:arch} have different meanings. The
\emph{enc-cut} is the concatenated visible encoder state and measures representation quality for
observed points. The \emph{emb-cut} is the learned $1232\!\to\!192$ map that prepares those features
for prediction. The \emph{pred-cut} is the predictor output at a queried, unobserved location. The
completion experiments in this paper use pred-cut:
the target object has been removed, so the readout cannot pool a visible feature from it. Encoder
semantic segmentation is retained only as a calibration of the representation, not as evidence for
latent completion.

\subsection{Occupied-region masking and deep supervision}

Masks are contiguous cubic blocks sampled over \emph{occupied} 3D cells
(Appendix Figure~\ref{fig:masking}). Restricting the mask
universe to occupied regions avoids spending the objective on trivially empty volume. During a
five-percent warmup, block size increases from $0.1$ to $0.4\,$m and the masked fraction from 0.3 to
0.7. The final loss balances prediction across all five encoder stages, so both local geometric state
and scene-level state supervise the predictor.

We pretrain on the 1,201 ScanNet training scenes~\citep{scannet} for 100 epochs with AdamW, a base
learning rate of 0.002, one-cycle scheduling, weight decay from 0.04 to 0.2, and gradient clipping at
1.0. Training uses random spatial and color augmentation after per-scene centering. The primary
evaluation checkpoint is epoch 95, fixed before endpoint scoring.

\section{Querying Missing Entity State}

The pretraining loss (Eq.~\ref{eq:loss}) says that a target latent is predictable; it does not say what the completed
state contains or why a downstream readout succeeds. We organize evaluation around four questions:
is the signal nontrivial, does the learned prediction path matter, does the output depend on the
correct scene, and how does the completed content support a structural decision?

\subsection{Deletion-first prediction}

For a target object $o$, we use its annotated geometry only to identify the points to delete and its
centroid, then remove all of its points \emph{before} grid sampling, centering, and encoding. The remaining
scene is processed once; visible anchor objects, when required by a relation item, are pooled from
this same target-deleted pass. We insert a fixed template of 32 points on a sphere of radius
$0.15\,$m at the oracle target centroid. The template is identical for every object, so query shape,
extent, and orientation cannot reveal its class. If a query token collides with a visible $2\,$cm
cell, the visible cell takes precedence.

The predictor (Eq.~\ref{eq:predict}) outputs one 1232-dimensional latent per
surviving query point, which we average to
obtain the imputed object state $\widehat{\mathbf z}_o$. The identity probe consumes this vector
alone. The support probe concatenates it with ordered pooled features for the visible anchor slots;
missing slots are padded consistently. Probes never receive target points or target box dimensions.

The oracle centroid intentionally separates \emph{state completion at a known location} from object
discovery. Localization from raw observations is an important but different problem. Here we ask what
the predictor infers once a location has been queried.

\subsection{Tests of learned completion and composition}

\paragraph{Signal beyond shortcut floors.}
For identity, we compare against a majority prior, a centroid-only MLP, three fully random models,
and shuffled-label probes. The identity gap is computed against the replicate-wise maximum applicable
floor. For support, coordinates, class pairs, fully random models, and shuffled labels are informative
marginal diagnostics, but not decisive controls because support depends on identity and geometry
jointly.

\paragraph{Learned predictive pathway.}
We keep the pretrained encoder fixed and replace its embedding map, predictor, and mask token with
independently randomized parameters. Every arm receives the same query, probe family, and selection
protocol; three grafts are evaluated per trained checkpoint. The trained-minus-graft contrast
measures the material contribution of predictor training and encoder--predictor co-adaptation. It
does not attribute the full completed state to the predictor alone.

\paragraph{Correct scene context.}
We replace the target's visible context with a preassigned cross-scene donor selected without labels.
The fixed query and clean trained probe are reused unchanged. In the identity endpoint, clean and
donor arms use the same union of recipient and donor deletion boxes, recipient-derived coordinate
frame, token count, and collision-pruned query, equalizing residual hole geometry. Support donors are
matched by visible token count and scene extent while recipient query bytes and anchor slots remain
fixed. Any decrease therefore measures dependence on the original surrounding scene under the same
readout, rather than adaptation of a new head. A zero-context arm is a secondary identity diagnostic.

\paragraph{Identity--geometry composition.}
The \emph{joint oracle} receives one-hot target and anchor classes with raw target--anchor centroids.
It tests task susceptibility: target class is unavailable to \method{} at inference, but a matching
score would show that support can be computed from compact entity and geometric inputs. The
\emph{predicted-identity control} replaces oracle target class with a 21-way posterior decoded from
the imputed target latent; anchor class and coordinates remain fixed. Identity decoders use five-fold
scene cross-fitting and are selected without held-out support labels. This tests whether information
actually completed by the model reproduces the full-latent readout.

For support, predictor and context interventions are also evaluated through this identity channel.
Differences between nonlinear AP contrasts are descriptive, not additive causal decompositions.
Appendix~\ref{sec:endpoint-details} gives the complete estimands
(Eq.~\ref{eq:estimands}) and donor construction.

\subsection{Endpoints}

\paragraph{External semantic identity.}
ARKitScenes provides mobile RGB-D scans and manually annotated 3D oriented boxes
\citep{arkitscenes} (Table~\ref{tab:endpoints}). We retain 17 official furniture classes and every box containing at least 40
occupied $2\,$cm cells. The official Training scenes are divided scene-disjointly into 42,725
fit-training and 7,899 fit-validation objects. The official Validation split is held out for one
evaluation and contains 5,953 objects from 549 scenes. Class frequency ranges from 25 to 2,255
objects, so the primary metric is 17-class macro top-1 accuracy. \method{} is pretrained only on
ScanNet; no ARKitScenes scene enters its encoder or predictor training.

\paragraph{Coarse support structure.}
Sr3D templates sharing a scene, target, and canonical anchor slots are aggregated into multilabel
relation items~\citep{referit3d}, avoiding contradictory single labels when an item bears multiple
relations. We evaluate binary support-family detection over 8,570 held-out target--anchor items from
255 scenes, of which 152 are support-positive. Average precision (AP) is the primary metric because
of this imbalance. The split is scene-disjoint, and validation AP selects the probe epoch without
access to held-out labels. A preliminary coordinate-only screen made the remaining coarse families
unsuitable for attribution, so they are excluded rather than averaged into a favorable relation
score. The endpoint does not test supporting-versus-supported-by direction.

\begin{table}[htbp]
  \centering
  \caption{\textbf{Two endpoints expose the content and composition of predicted scene state.} Both
  remove the target before feature extraction and use the same fixed query geometry. ARKitScenes is
  external to pretraining and tests completed entity identity; Sr3D is ScanNet-derived and tests how
  that entity state combines with geometry in a downstream structural decision.}
  \label{tab:endpoints}
  \small
  \setlength{\tabcolsep}{5pt}
  \begin{tabular*}{\linewidth}{@{\extracolsep{\fill}}l l l@{}}
    \toprule
    & ARKitScenes identity & Sr3D support family \\
    \midrule
    Held-out set & 5,953 objects; 549 scenes; 17 classes & 8,570 pairs; 255 scenes; 152 positives \\
    Metric & macro top-1 accuracy & average precision \\
    Readout & imputed target latent & imputed target + visible anchor slots \\
    Key control & fully random model (20.95) & identity + xyz oracle (43.63) \\
    Context test & matched donor scene & full + identity-channel donor \\
    Supported inference & entity identity & support from identity + geometry \\
    \bottomrule
  \end{tabular*}
\end{table}

\subsection{Probes and uncertainty}

All readouts are two-layer MLPs trained on frozen features, with a 256-dimensional hidden layer,
GELU, dropout 0.1, and inverse-square-root class weighting. They are trained with AdamW for at most
50 epochs. We use three
pretrained \method{} checkpoints and three head seeds. The identity endpoint also evaluates three
checkpoints of the canonical recipe as a replication; the support predictor contrast averages three
random graft seeds within each checkpoint.

Confidence intervals use 100,000 paired scene-cluster bootstrap replicates. Each resamples scenes,
recomputes every head/graft/checkpoint realization, and then performs the prespecified nested
averaging. The intervals account for within-scene dependence but remain conditional on the three
pretrained checkpoints; they are not population intervals over training runs.

\section{Results}

Figure~\ref{fig:overview} makes the deletion-first query protocol concrete. Table~\ref{tab:main}
reports the claim-bearing endpoint estimates. The retained predictor completes entity identity across
two training recipes; the structural endpoint then shows how that entity state combines with geometry.
Figure~\ref{fig:relationevidence} visualizes this composition. The full unsubtracted control ledger is
retained in Appendix Table~\ref{tab:rawmain}.

\begin{table}[htbp]
  \centering
  \caption{\textbf{Completed entity state and its structural composition.} Panel A establishes
  semantic completion at a known location. Panel B tests whether support requires information beyond
  identity and geometry. Effects are metric points with 95\% paired scene-bootstrap intervals.
  Context-arm scores use separately paired geometry and therefore need not equal primary scores.
  Differences between AP contrasts are descriptive, not additive causal decompositions.}
  \label{tab:main}
  \small
  \setlength{\tabcolsep}{5pt}
  \begin{tabular*}{\linewidth}{@{\extracolsep{\fill}}l r r@{}}
    \toprule
    Contrast & Compared scores & Effect [95\% CI] \\
    \midrule
    \multicolumn{3}{@{}l}{\textbf{A. Entity completion} \quad \textit{ARKitScenes mAcc}}\\[1pt]
    trained $-$ strongest floor & 43.13 vs. 20.95 &
      \textbf{+22.18} [20.83, 23.53] \\
    trained $-$ randomized predictor & 43.13 vs. 33.35 &
      \textbf{+9.78} [8.55, 11.03] \\
    correct $-$ donor context & 32.94 vs. 10.95 &
      \textbf{+21.98} [20.67, 23.30] \\
    \midrule
    \multicolumn{3}{@{}l}{\textbf{B. Structural composition} \quad \textit{Sr3D support AP}}\\[1pt]
    full $-$ joint identity + geometry & 41.15 vs. 43.63 &
      $-2.48$ [$-11.09$, 5.98] \\
    full $-$ predicted identity + geometry & 41.15 vs. 39.37 &
      +1.78 [$-5.11$, 9.06] \\
    context effect: full latent & 41.14 vs. 23.54 &
      \textbf{+17.60} [12.08, 23.05] \\
    context effect: identity channel & --- &
      \textbf{+15.57} [10.36, 20.83] \\
    context excess (full $-$ identity) & --- &
      +2.03 [$-3.92$, 7.95] \\
    \bottomrule
  \end{tabular*}
\end{table}

\subsection{The predicted state identifies an unseen object}

On the held-out ARKitScenes split, \method{} reaches 43.13\% macro accuracy from the imputed latent
of a deleted object. The strongest shortcut is the fully random model at 20.95\%, slightly above the
centroid-only floor at 19.23\%; majority and shuffled-label probes are both approximately 5.88\%.
The primary trained-minus-maximum-floor effect is therefore \textbf{+22.18 points}, with a 95\%
interval of [20.83, 23.53] (Table~\ref{tab:main}). The encoder and predictor remain frozen after
ScanNet pretraining; only the MLP readout is fit on ARKitScenes Training rooms.

This behavior is not confined to the per-stage deep-supervision recipe. A second SR-JEPA recipe
without per-stage loss balancing reaches 44.61\% macro accuracy across its three checkpoints. We use
the prespecified primary model for all controlled contrasts, but this replication shows that identity
completion is stable across the two trained recipes tested rather than arising from one selected
checkpoint.

Retaining the encoder while randomizing the predictor lowers macro accuracy from 43.13\% to 33.35\%,
a \textbf{+9.78}-point predictor training/co-adaptation effect [8.55, 11.03]. Random features can
still expose some scene and location regularity to a trained MLP; they do not account for the full
imputed state.

The decisive contrast replaces the correct scene with a matched donor while preserving the fixed
query and clean head. Accuracy falls from 32.94\% to 10.95\%, a \textbf{+21.98}-point context effect
[20.67, 23.30]. With no context it falls further to 6.05\%. Because the readout is unchanged, this
cannot be explained by fitting a new classifier to a different feature distribution. The semantic
identity decoded at the query depends on the target's actual surrounding scene.

Class-wise gains span multiple categories rather than only the dominant cabinet class
(Appendix Figure~\ref{fig:identityclass} and Table~\ref{tab:perclass}). Individual rare-class bars remain descriptive; inference
rests on the scene-bootstrap macro effects.

\subsection{Support decisions compose completed identity with geometry}

The imputed state is useful for a downstream structural decision: concatenating the predicted target
latent with visible anchor features yields 41.15 support AP. Marginal controls initially appear to
leave a large gap---raw coordinates reach 29.03 AP, class pairs 10.38 AP, fully random models 12.74
AP, and shuffled labels 2.29 AP---but support is an interaction. Geometry describes vertical/contact
configuration; identity determines whether the participating object roles make that configuration a
support event.

The joint controls expose this interaction. Ground-truth target and anchor identities plus their raw
coordinates reach 43.63 AP, 2.48 points above the full latent; the paired interval for
$\Delta_{\mathrm{joint}}$ is [$-11.09$, 5.98]. Replacing ground-truth target class with the identity
posterior decoded from \method{} still reaches 39.37 AP. The full latent exceeds
this model-level composition control by 1.78 points [$-5.11$, 9.06]. Thus completed identity and
geometry reproduce most observed support utility; the present endpoint does not resolve additional
support information in the full latent.

The interventions remain real but admit the same explanation. Randomizing the predictor reduces the
full support readout by 8.79 AP [4.38, 13.81]; through the identity channel alone, it reduces support
by 12.89 AP [7.68, 17.85]. Replacing correct context with a donor reduces the full readout by 17.60 AP
[12.08, 23.05] and the identity-channel readout by 15.57 AP [10.36, 20.83]
(Appendix Table~\ref{tab:channels}). The descriptive excess
of the full context effect is 2.03 AP [$-3.92$, 7.95]. Thus context and predictor training matter,
but their structural utility is largely carried by the entity identity inferred at the missing
location.

\begin{figure}[htbp]
  \centering
  \includegraphics[width=\linewidth]{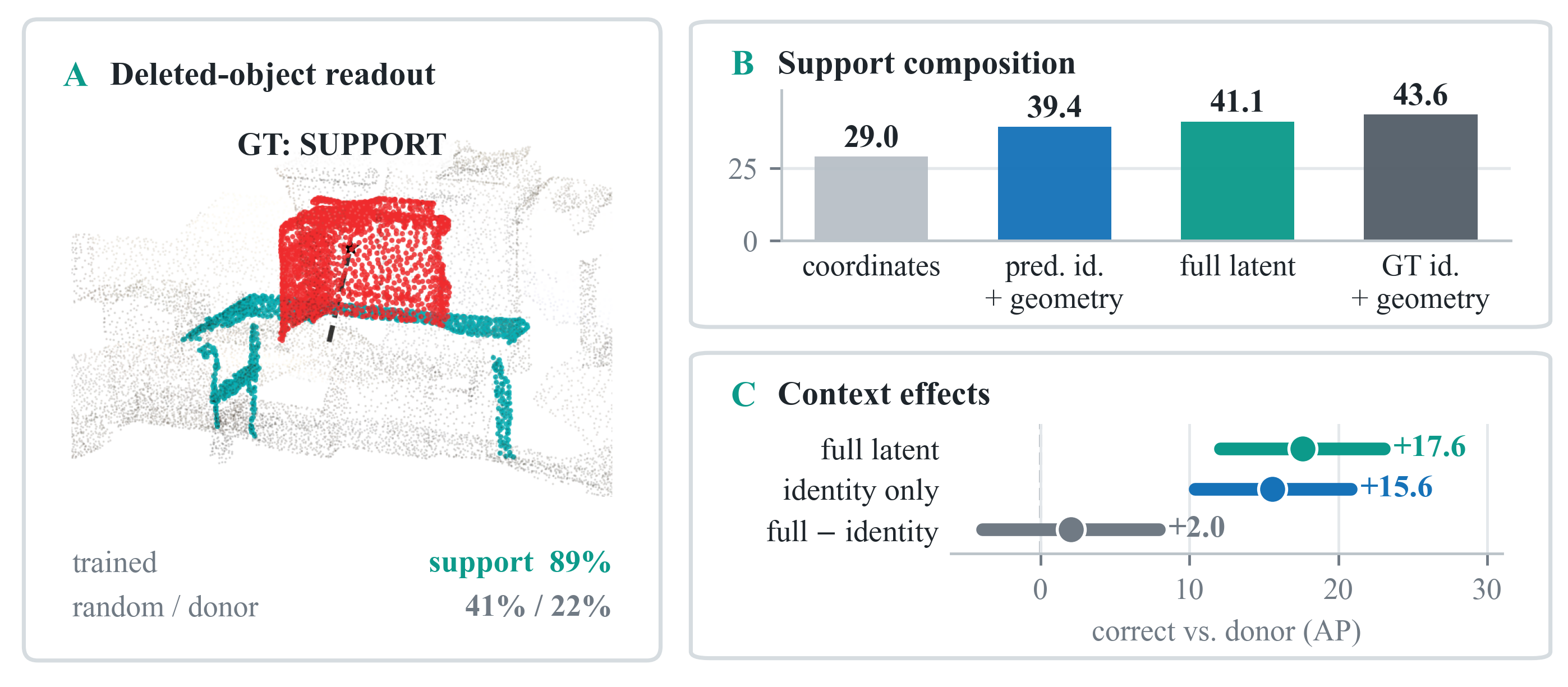}
  \caption{\textbf{Completed entity state combines with geometry to support structural decisions.}
  \textbf{(A)} An outcome-blind fixed-hash item from the final endpoint. Red target points are restored
  only for the reader; the cyan anchor remains visible. Rows give the trained vote followed by the
  randomized-predictor and donor-context votes. The example illustrates the downstream readout;
  Panels B--C test its composition. \textbf{(B)} Full predicted latents substantially exceed
  geometry alone, while joint identity--geometry controls reproduce most of that utility.
  \textbf{(C)} Most of the
  correct-context effect is reproduced when the support head receives only predicted target identity,
  anchor identity, and coordinates. Intervals are 95\% paired scene bootstraps over all 8,570 items.}
  \label{fig:relationevidence}
\end{figure}

\subsection{One predictive state, two levels of computation}

The results identify a clean division of labor. \method{} uses the surrounding scene to estimate
\emph{what entity belongs at an unobserved location}. Metric coordinates then provide \emph{where},
and a lightweight downstream function can derive support. A useful state need not contain a dedicated
token for every relation; entity state and pose form a compositional basis from which relations can be
computed. The present evidence does not support the stronger claim that support is represented
independently of those constituents. This factorization is itself the representation-level finding:
the completed entity state provides a compact basis from which a downstream computation derives a
structural relation.

\section{Discussion and Limitations}

\paragraph{The predictor as a state interface.}
Our main result concerns behavior, not an encoder leaderboard. Complete-object deletion gives the
frozen encoder--predictor pathway a concrete test: produce state at a location where entity evidence
is absent. Randomized-predictor and donor-context interventions show that the output depends on both
the learned prediction path and the correct scene. The joint controls show that its downstream utility
comes from completed entity content combined with geometry. Because both sides of the
predictive objective are functions of the point scene, this state is self-contained when cameras,
language, or a fixed 2D teacher are unavailable. Lifted image features can provide stronger semantic
supervision~\citep{locate3d}; our result is that native 3D prediction already produces usable missing
entity state, not that external supervision is undesirable.

\paragraph{Why the state composes.}
The occupied-region objective regresses pointwise EMA features with $L_1$ loss. It contains no object
slot, target--anchor pair, relation token, or term that requires support to be represented
independently. The loss can therefore be minimized by inferring an entity prototype compatible with
the query and context; a downstream function can then combine identity with metric geometry. The
full-object endpoint shows that region-level training generalizes to missing entities, and the joint
controls are consistent with this factorization. Object-level masking or interaction targets, as in
C-JEPA~\citep{cjepa}, offer a direct alternative when interactions must be explicit in the state.

\paragraph{Scope of completion.}
The supplied centroid isolates state completion from discovery: \method{} is not asked to detect that
an entity is missing or to localize it. The fixed query removes shape, but a single deletion can leave
boundary cues; the paired donor construction equalizes this geometry and therefore provides the
stronger context attribution. The current $L_1$ objective also returns one deterministic latent rather
than multiple plausible hidden states. Finally, the evidence establishes use of visible surroundings,
not room-scale dependence. A coverage-limited post-hoc diagnostic found no resolved incremental
identity benefit beyond $1.5\,$m, while the $3\,$m arm was unsupported by the token cap
(Table~\ref{tab:spatialextent}). Spatial reach therefore remains open.

\paragraph{Structural and empirical scope.}
The Sr3D endpoint contains 152 support-positive items, does not distinguish supporting from
supported-by, and shares the ScanNet source corpus with pretraining. It characterizes composition,
not general spatial reasoning, physical stability, or action-induced change. Three of four sequential
primary launches completed; one diverged with non-finite loss at epoch 19 under an automated rule.
Intervals are conditional on the completed checkpoints, and the support residual is not consistently
positive across them (Table~\ref{tab:checkpoint}). The study also leaves scaling open. These
experiments characterize this JEPA objective and evaluation interface; they do not
establish that latent prediction is necessary for contextual completion relative to matched
reconstruction, contrastive, or alternative predictive objectives. A non-matched ScanNet calibration
places the visible encoder above a random PT-v3 but below large-scale Sonata (Appendix
Table~\ref{tab:semseg}); encoder recognition and predictive completion are distinct capabilities.

\paragraph{From completion to transition.}
Deleting an object changes the observation, not the physical world. The present results therefore
establish static state inference under partial observability, not dynamics or planning. The next test
is an action-conditioned predictor over the same native 3D state. That claim would require transition,
rollout, and control metrics rather than a flexible probe. The evaluation principle is the same:
report both what the encoder recognizes and what the predictive pathway can complete, transform, or
roll forward.

\section{Conclusion}

We presented \method, a point-native JEPA whose original frozen predictive pathway can be queried for
state at a supplied, unobserved 3D location. Under complete-object deletion, the imputed latent
recovers semantic entity identity on external ARKitScenes data. Shortcut floors establish that the
signal is nontrivial, while predictor and context interventions show that completion depends on both
the learned prediction path and the correct surrounding scene. On Sr3D, the same state supports a
structural decision; joint controls reveal its organization. Completed target identity, anchor
identity, and metric geometry reproduce most full-latent support utility, exposing a compositional
division between inferred entity content and downstream geometric computation.

For \method{}, the predictor is therefore more than transient pretraining machinery: it is an
operational interface for present-state inference under partial observability. The current interface
is static, deterministic, and queried at a known location, but it provides a concrete test for
predictive representations: report both what the encoder recognizes and what the learned predictive
pathway completes when evidence is absent. Extending that interface to discovery,
multiple hypotheses, and action-conditioned transition is the next problem and requires new evidence.

\FloatBarrier


\clearpage
\appendix

\section{Expanded Method Positioning}

\begin{table}[htbp]
  \centering
  \caption{\textbf{Positioning among closely related 3D predictive methods.} ``Native'' means that
  the target is learned from 3D input without lifted 2D foundation features. MSP overlaps in
  scene-level masked deep-feature prediction, and AD-L-JEPA analyzes predicted occupancy
  qualitatively. The last column asks the narrower question central here: is the retained predictor
  quantitatively evaluated after complete-object deletion with predictor and context interventions?
  The table defines a design conjunction, not a first/only claim.}
  \label{tab:landscape}
  \small
  \setlength{\tabcolsep}{3.8pt}
  \begin{tabular*}{\linewidth}{@{\extracolsep{\fill}}l l l l c@{}}
    \toprule
    Method & Scale & Target source & Pred. eval. & Deletion controls \\
    \midrule
    MSP~\citep{msp} & scene & shape + deep feature & no & no \\
    Point-JEPA~\citep{pointjepa} & object & native 3D EMA & no & no \\
    3D-JEPA (Hu)~\citep{hu3djepa} & object & target-block latent & no & no \\
    Locate 3D~\citep{locate3d} & indoor scene & 2D priors + EMA & no & no \\
    AD-L-JEPA~\citep{adljepa} & LiDAR / BEV & native LiDAR latent & qualitative & no \\
    AD-LiST-JEPA~\citep{adlistjepa} & LiDAR sequence & future latent & no & no \\
    POMA-3D~\citep{poma3d} & room map & 2D-prior latent & no & no \\
    \method{} & indoor scene & native 3D EMA & quantitative & \textbf{yes} \\
    \bottomrule
  \end{tabular*}
\end{table}

\section{Reproducibility Details}

\paragraph{Architecture.}
The encoder is PT-v3m2 with $2\,$cm voxels and receives coordinates, RGB, and surface normals.
Stage depths are $(3,3,3,12,3)$, widths are
$(48,96,192,384,512)$, and stage up-cast concatenation yields 1232 dimensions.
The predictor has width 192, depth 6, and 6 heads. Its input and output projections are
$1232\!\to\!192$ and $192\!\to\!1232$, respectively. Serialization order is fixed during evaluation.

\paragraph{Optimization.}
Cubic occupied-region masks use size $0.1\!\to\!0.4\,$m and ratio $0.3\!\to\!0.7$. The loss is
five-stage $L_1$ latent prediction plus the visible-token predictive term. EMA momentum is
$0.994\!\to\!1.0$. We train for 100 epochs with batch size 2, bfloat16, AdamW, base learning rate
0.002, weight decay $0.04\!\to\!0.2$, layer-wise decay 0.9, OneCycle scheduling, and gradient clip
1.0. Primary checkpoints use epoch 95 and seeds 239, 240, and 242. Four seeds were launched in
sequence; seed 241 encountered a non-finite-loss divergence at epoch 19 and produced no endpoint
value. No completed run was filtered by downstream performance.

\begin{figure}[htbp]
  \centering
  \includegraphics[width=\linewidth]{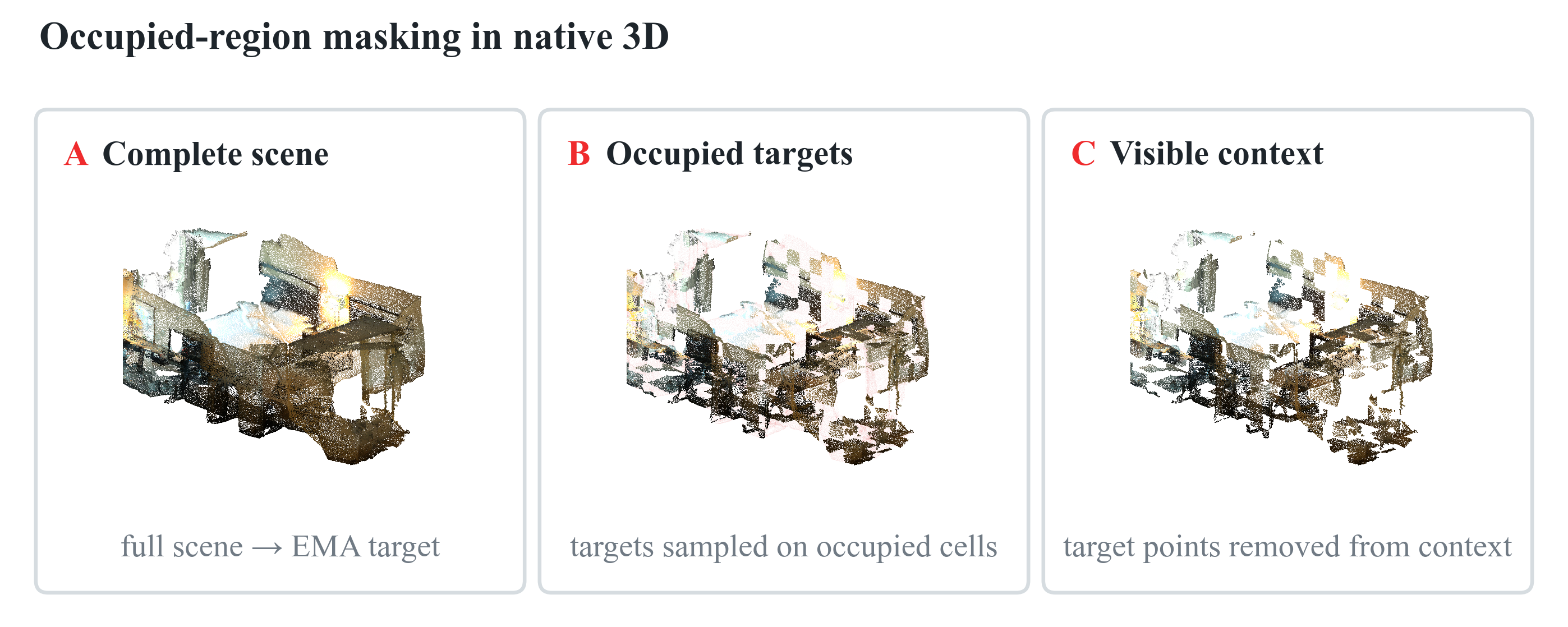}
  \caption{\textbf{Occupied-region masking remains native to the observed 3D scene.}
  The EMA target encoder receives the complete point scene \textbf{(A)}; cubic target regions are
  sampled only from occupied cells \textbf{(B)}; and those target points are absent from the context
  encoder input \textbf{(C)}. The objective therefore predicts occupied latent state rather than
  reconstructing empty volume.}
  \label{fig:masking}
\end{figure}

\paragraph{Preprocessing.}
Training and evaluation grid-sample at $0.02\,$m and apply the same color normalization and
\texttt{CenterShift(apply\_z=True)}. Training additionally applies random scale, rotation, flip,
jitter, elastic distortion, and color perturbation. Endpoint preparation deletes the target before
grid sampling and computes the scene transform from visible points only.

\paragraph{Probe fitting.}
Heads use a 256-dimensional hidden layer, GELU, dropout 0.1, AdamW learning rate $10^{-3}$ and weight
decay $10^{-4}$, OneCycle scheduling, and inverse-square-root class weighting. Head seeds are 0, 1,
and 2. Epoch selection uses only scene-disjoint fit-validation data. Context interventions reuse the
clean selected head byte-for-byte.

\section{Endpoint Protocol Details}
\label{sec:endpoint-details}

\paragraph{Target deletion and query construction.}
For each endpoint item, target membership is resolved from the dataset annotation before any model
input is built. All target points are removed, the remaining scene is centered and grid-sampled, and
only then are query coordinates inserted. The query is a deterministic 32-point sphere of radius
$0.15\,$m centered on the annotated target centroid; class, target dimensions, yaw, and target point
samples do not affect its bytes. A visible voxel wins any coordinate collision, preventing a query
token from replacing observed context. Query survival counts and serialized coordinates are checked
across paired arms.

\paragraph{ARKitScenes split and context intervention.}
The official Training partition supplies probe-fit data and is divided by scene into 42,725 training
and 7,899 validation objects. Official Validation is untouched until final scoring and contains 5,953
eligible objects in 549 scenes. For the donor intervention, each recipient is paired with a donor
scene without using semantic labels. Both clean and donor conditions remove the union of the
recipient and donor target boxes, use the recipient coordinate frame, enforce the same visible-token
budget, and receive the same collision-pruned query. This ``union-hole'' construction prevents a
classifier from identifying the arm through a recipient-shaped versus donor-shaped deletion boundary.
Donors form a deterministic cross-scene bijection: items are ranked by visible-token-count and
visible-bounding-box-diagonal deciles, scenes are ordered by block size and identifier, and donor
blocks are rotated and matched monotonically with item identifier as the final tie-break. The head
selected on the clean fit-validation partition is applied unchanged to both conditions.

\paragraph{Sr3D multilabel repair and donor construction.}
Original Sr3D descriptions can assign more than one coarse relation to the same scene--target--anchor
configuration. We collapse templates sharing the scene, target, and canonical ordered anchor slots
into one multilabel item before splitting. Probe train, validation, and held-out partitions are
scene-disjoint. The final endpoint reports only the support family because the preregistered raw-
coordinate screen made other coarse families non-attributable. Donor scenes are selected without
relation labels and matched on visible token count and spatial extent. Recipient query bytes and
anchor feature slots remain fixed, and the clean-selected head is reused for donor scoring. For
two-anchor items, slots follow ascending instance identifier. The deterministic donor map ranks token
count and bounding-box diagonal into deciles, permutes items with a fixed protocol seed, and selects
the nonzero cyclic shift that lexicographically minimizes total token-decile distance, total
diagonal-decile distance, and shift while rejecting self- and same-scene pairs.

\paragraph{Joint identity--geometry controls.}
The oracle control concatenates one-hot target and anchor identities with raw target--anchor
centroids. The model-level control replaces target identity with a 21-way (ScanNet20 plus unknown)
posterior decoded from the imputed target latent, while preserving oracle anchor identity and
coordinates. Identity decoders use five-fold scene-cross-fitting on probe-training scenes; their
epochs and the support-head epoch are selected on scene-disjoint validation data. Both controls use
the same 256-hidden-unit MLP family as the main support readout. Predictor-graft and donor-context
arms reuse the clean-selected identity and support heads, changing only the target posterior.

\paragraph{Random-model and predictor-graft controls.}
The fully random floor instantiates the complete encoder, embedding map, predictor, and mask token
without self-supervised training; it is not a low-dimensional noise vector. The predictor graft keeps
the trained encoder but replaces the embedding/predictor path and mask token with an independently
initialized counterpart of the same architecture. Three random-model realizations and, where
applicable, three grafts are nested within every checkpoint and head seed. Thus
$\Delta_{\mathrm{pred}}$ measures the value of predictor training and encoder--predictor
co-adaptation; it does not claim that the encoder contributes nothing to the graft score.

\paragraph{Selection and uncertainty.}
No held-out label participates in model extraction, preprocessing, probe epoch selection, or donor
assignment. The declared joint controls use held-out target/anchor class labels only as final
comparator inputs after their architectures and epochs are fixed; the predicted-identity control
replaces target class but retains anchor class. Held-out support labels are opened only by the final
scorer after prediction bundles and manifests are fixed. Each of 100,000 accepted bootstrap replicates
resamples scenes with replacement, preserving all objects/items and paired arms within a sampled
scene. Metrics are recomputed per realization before nesting is averaged. This design respects scene
clustering and pairing, while the resulting intervals remain conditional on the three completed
pretraining checkpoints rather than estimating variability over an unlimited population of training
runs.

\paragraph{Reported estimands.}
Let $\mathcal{M}$ denote macro accuracy or support AP as appropriate, $T$ the trained system, $F_k$
an applicable identity floor, $G$ the randomized-predictor graft, $C,D$ the correct- and donor-context
arms, $J$ the joint identity--geometry oracle, and $P$ the predicted-identity composition control. We
report
\begin{equation}
\begin{aligned}
  \Delta_{\mathrm{floor}} &= \mathcal{M}(T)-\max_k \mathcal{M}(F_k), \\
  \Delta_{\mathrm{pred}}  &= \mathcal{M}(T)-\mathcal{M}(G), \\
  \Delta_{\mathrm{ctx}}   &= \mathcal{M}(C;h^*)-\mathcal{M}(D;h^*), \\
  \Delta_{\mathrm{joint}} &= \mathcal{M}(T)-\mathcal{M}(J), \\
  \Delta_{\mathrm{comp}}  &= \mathcal{M}(T)-\mathcal{M}(P).
\end{aligned}
\label{eq:estimands}
\end{equation}
The maximum identity floor is recomputed within each bootstrap replicate. Context effects use the
clean-selected head $h^*$, and their separately paired geometry means the clean score need not equal
the primary trained score.

\section{Detailed Endpoint Results}

\begin{figure}[htbp]
  \centering
  \includegraphics[width=\linewidth]{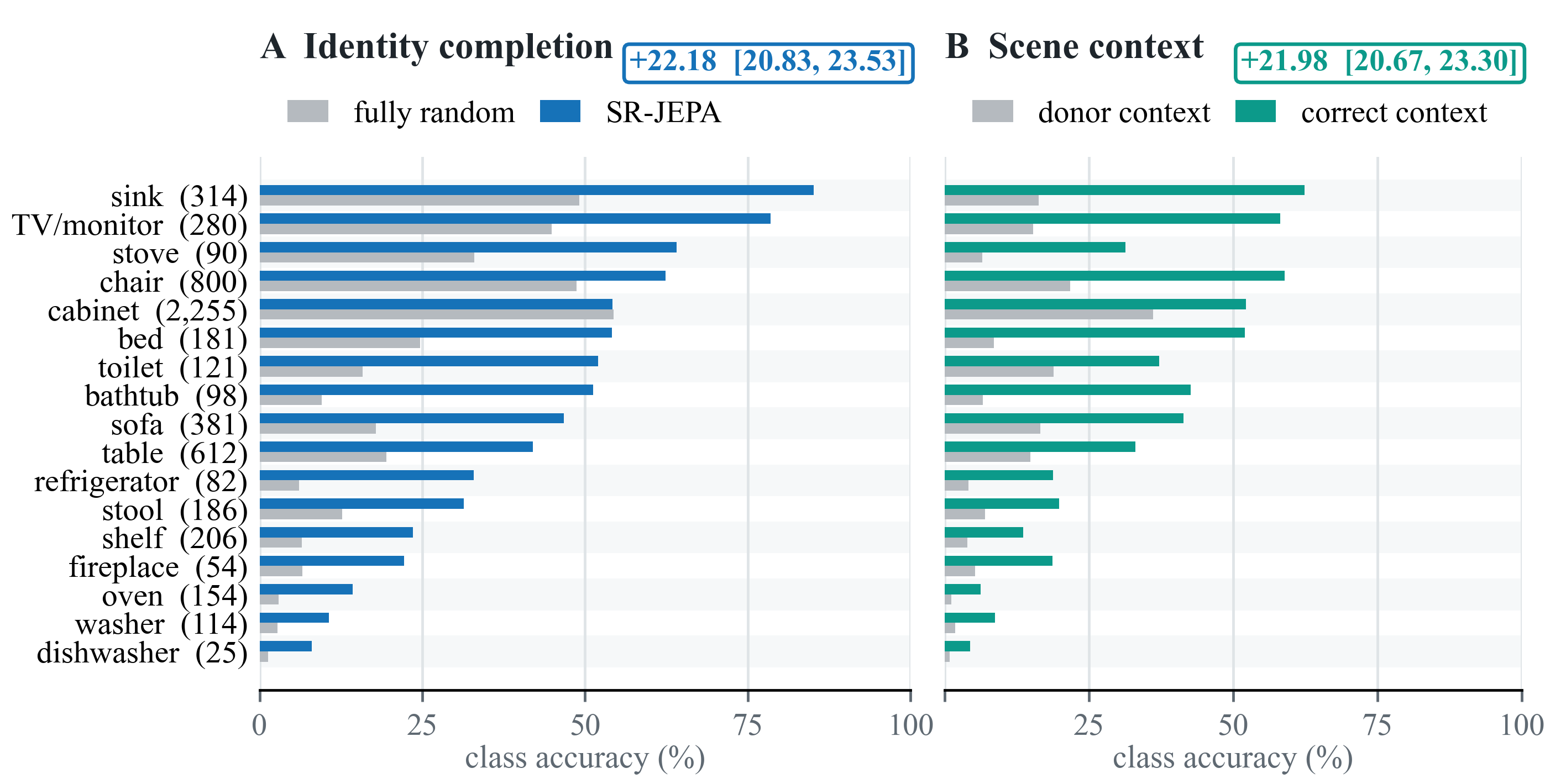}
  \caption{\textbf{Identity completion is distributed across the official ARKitScenes classes.}
  Grouped bars show descriptive class accuracies; inset values are the claim-bearing macro effects
  with 95\% scene-bootstrap intervals. Panels use different paired deletion geometries and should not
  be subtracted. Parentheses give held-out counts; individual class bars have no intervals and are
  especially uncertain for rare classes.}
  \label{fig:identityclass}
\end{figure}

\begin{table}[htbp]
  \centering
  \caption{\textbf{Frozen-encoder semantic segmentation is a calibration, not the claimed axis.}
  ScanNet validation mIoU uses the same MLP-probe harness and three head seeds. Sonata-HF uses a
  similar PT-v3 encoder but roughly 140,000 point clouds from seven datasets, including ScanNet
  geometry; \method{} uses 1,201 ScanNet training scenes. The comparison is therefore a practical
  boundary, not an objective- or exposure-matched ranking.}
  \label{tab:semseg}
  \small
  \begin{tabular}{lccc}
    \toprule
    Frozen encoder & Approx. pretraining scenes & Retained predictor & ScanNet mIoU \\
    \midrule
    Random PT-v3 & 0 & no & 22.9 \\
    \method{} & 1,201 & yes & 39.0 \\
    Sonata-HF~\citep{sonata} & $\sim$140,000 & no & \textbf{72.3} \\
    \bottomrule
  \end{tabular}
\end{table}
\FloatBarrier

\begin{table}[htbp]
  \centering
  \caption{\textbf{Raw endpoint scores and control arms (\%).} Identity is 17-class macro
  accuracy; structure is support-family AP. ``Geometry'' is a centroid-only identity probe or a raw
  target--anchor coordinate probe. Fully random uses the complete untrained encoder--predictor
  architecture. Context rows reuse the selected clean head; their separately paired geometry is why
  their clean score need not equal the primary row.}
  \label{tab:rawmain}
  \small
  \setlength{\tabcolsep}{5pt}
  \begin{tabular*}{\linewidth}{@{\extracolsep{\fill}}lccp{0.38\linewidth}@{}}
    \toprule
    Arm & Identity mAcc & Support AP & Information changed \\
    \midrule
    \method{} primary & \textbf{43.13} & \textbf{41.15} & trained encoder, predictor, and correct scene \\
    Geometry-only & 19.23 & 29.03 & target location / raw target--anchor coordinates \\
    Class or majority prior & 5.88 & 10.38 & label frequency / visible class pair \\
    Joint GT identity + geometry & --- & \textbf{43.63} & task-susceptibility oracle \\
    Predicted identity + geometry & --- & 39.37 & model-level composition control \\
    Fully random model & 20.95 & 12.74 & no self-supervised training \\
    Shuffled-label probe & 5.87 & 2.29 & probe selectivity control \\
    Random-predictor graft & 33.35 & 32.36 & trained encoder; random predictor and mask token \\
    Correct context, fixed head & 32.94 & 41.14 & donor-paired clean geometry \\
    Donor context, fixed head & 10.95 & 23.54 & wrong scene; same query and readout \\
    Zero context, fixed head & 6.05 & --- & query without visible scene (secondary) \\
    \bottomrule
  \end{tabular*}
\end{table}

\begin{table}[htbp]
  \centering
  \caption{\textbf{Support interventions through the full latent and identity channel.} Entries are
  AP effects with 95\% paired scene-bootstrap intervals. The final column subtracts two nonlinear AP
  contrasts and is descriptive rather than an additive causal decomposition.}
  \label{tab:channels}
  \small
  \setlength{\tabcolsep}{4pt}
  \begin{tabular*}{\linewidth}{@{\extracolsep{\fill}}lrrr@{}}
    \toprule
    Intervention & full latent & identity channel & full $-$ identity \\
    \midrule
    trained $-$ random predictor
      & +8.79 [4.38, 13.81] & +12.89 [7.68, 17.85] & $-4.10$ [$-9.89$, 2.48] \\
    correct $-$ donor context
      & +17.60 [12.08, 23.05] & +15.57 [10.36, 20.83] & +2.03 [$-3.92$, 7.95] \\
    \bottomrule
  \end{tabular*}
\end{table}

\begin{table}[htbp]
  \centering
  \caption{\textbf{ARKitScenes class-wise accuracy (\%).} The first four score columns use the
  primary deletion geometry. Correct and donor context use their separately matched union-hole
  geometry and the same frozen head. The aggregate claim is macro-averaged and bootstrap-tested;
  individual class rows are descriptive, particularly when $N$ is small.}
  \label{tab:perclass}
  \small
  \setlength{\tabcolsep}{5pt}
  \begin{tabular*}{\linewidth}{@{\extracolsep{\fill}}lrrrrr@{}}
    \toprule
    \multicolumn{6}{@{}l}{\textbf{A. Primary deletion geometry}}\\[2pt]
    Class & $N$ & \method{} & Centroid & Fully random & Random predictor \\
    \midrule
    cabinet      & 2,255 & 54.2 & 44.4 & 54.3 & 55.2 \\
    refrigerator & 82    & 32.9 & 0.0  & 6.1  & 16.7 \\
    shelf        & 206   & 23.6 & 1.0  & 6.5  & 12.8 \\
    stove        & 90    & 64.1 & 0.0  & 33.0 & 49.4 \\
    bed          & 181   & 54.1 & 0.0  & 24.7 & 45.4 \\
    sink         & 314   & 85.1 & 92.8 & 49.1 & 74.4 \\
    washer       & 114   & 10.6 & 0.0  & 2.7  & 4.5 \\
    toilet       & 121   & 52.0 & 0.0  & 15.8 & 39.3 \\
    bathtub      & 98    & 51.2 & 3.1  & 9.5  & 28.9 \\
    oven         & 154   & 14.3 & 0.0  & 2.9  & 9.0 \\
    dishwasher   & 25    & 8.0  & 0.0  & 1.3  & 2.1 \\
    fireplace    & 54    & 22.2 & 0.0  & 6.6  & 14.7 \\
    stool        & 186   & 31.4 & 40.9 & 12.7 & 21.5 \\
    chair        & 800   & 62.4 & 74.6 & 48.7 & 56.0 \\
    table        & 612   & 42.0 & 41.6 & 19.5 & 30.6 \\
    TV/monitor   & 280   & 78.5 & 27.7 & 44.9 & 64.6 \\
    sofa         & 381   & 46.7 & 0.8  & 17.9 & 41.7 \\
    \midrule
    Macro        & 5,953 & \textbf{43.13} & 19.23 & 20.95 & 33.35 \\
    \bottomrule
  \end{tabular*}

  \vspace{0.8em}
  \begin{tabular*}{\linewidth}{@{\extracolsep{\fill}}lrrr@{}}
    \toprule
    \multicolumn{4}{@{}l}{\textbf{B. Matched context intervention}}\\[2pt]
    Class & $N$ & Correct context & Donor context \\
    \midrule
    cabinet      & 2,255 & 52.2 & 36.1 \\
    refrigerator & 82    & 18.8 & 4.2 \\
    shelf        & 206   & 13.6 & 3.9 \\
    stove        & 90    & 31.4 & 6.5 \\
    bed          & 181   & 52.1 & 8.6 \\
    sink         & 314   & 62.4 & 16.3 \\
    washer       & 114   & 8.8  & 1.9 \\
    toilet       & 121   & 37.2 & 18.9 \\
    bathtub      & 98    & 42.6 & 6.7 \\
    oven         & 154   & 6.3  & 1.2 \\
    dishwasher   & 25    & 4.4  & 0.9 \\
    fireplace    & 54    & 18.7 & 5.3 \\
    stool        & 186   & 19.9 & 7.0 \\
    chair        & 800   & 58.9 & 21.8 \\
    table        & 612   & 33.1 & 14.9 \\
    TV/monitor   & 280   & 58.1 & 15.4 \\
    sofa         & 381   & 41.4 & 16.6 \\
    \midrule
    Macro        & 5,953 & 32.94 & 10.95 \\
    \bottomrule
  \end{tabular*}
\end{table}
\FloatBarrier

\begin{table}[htbp]
  \centering
  \caption{\textbf{Post-hoc spatial-reach diagnostic on ARKitScenes identity (mAcc, \%).}
  The local neighborhood ($\leq r$) and query are fixed while context beyond
  $r+0.04\,$m is either retained, replaced by count-matched donor context, or removed.
  Effects are paired scene-bootstrap intervals. The preregistered coverage gate requires at least
  1,000 objects, 100 scenes, and all 17 classes; failed gates make both rows inconclusive rather than
  negative. No model or readout is refit.}
  \label{tab:spatialextent}
  \small
  \setlength{\tabcolsep}{6pt}
  \begin{tabular*}{\linewidth}{@{\extracolsep{\fill}}lrrl@{}}
    \toprule
    \multicolumn{4}{@{}l}{\textbf{A. Coverage}}\\[2pt]
    Radius & Objects & Scenes & Validity \\
    \midrule
    $1.5\,$m & 442 & 268 & coverage fail \\
    $3.0\,$m & 0 & 0 & unsupported by retained tokens \\
    \bottomrule
  \end{tabular*}

  \vspace{0.65em}
  \begin{tabular*}{\linewidth}{@{\extracolsep{\fill}}lrrr@{}}
    \toprule
    \multicolumn{4}{@{}l}{\textbf{B. Reconstructed scores}}\\[2pt]
    Radius & local + correct far & local + donor far & local only \\
    \midrule
    $1.5\,$m & 30.19 & 30.24 & 31.02 \\
    $3.0\,$m & --- & --- & --- \\
    \bottomrule
  \end{tabular*}

  \vspace{0.65em}
  \begin{tabular*}{\linewidth}{@{\extracolsep{\fill}}lcc@{}}
    \toprule
    \multicolumn{3}{@{}l}{\textbf{C. Paired effects}}\\[2pt]
    Radius & correct $-$ donor [95\% CI] & correct $-$ local [95\% CI] \\
    \midrule
    $1.5\,$m & $-0.05$ [$-4.39$, 3.09] & $-0.83$ [$-3.21$, 1.74] \\
    $3.0\,$m & --- & --- \\
    \bottomrule
  \end{tabular*}
\end{table}

At $1.5\,$m all 17 classes remain represented and the reconstructed clean arm retains
1.01 times the original clean lift above the constant-class floor, but only 442 of 5,953 objects
have the required local and far token counts. At $3\,$m no object retains enough correct and donor
far tokens after nearest-token capping. The diagnostic therefore neither establishes a useful
non-local contribution nor licenses a definitive null.

\begin{table}[htbp]
  \centering
  \caption{\textbf{Checkpoint-level replication.} Identity columns are means over three head seeds.
  ``Single-level'' is the same JEPA recipe without per-stage loss balancing. Support columns compare
  the full latent with the joint oracle and predicted-identity control. Identity completion is stable;
  neither support residual is consistently positive. Launch 241 diverged before evaluation.}
  \label{tab:checkpoint}
  \small
  \begin{tabular*}{\linewidth}{@{\extracolsep{\fill}}lrrrr@{}}
    \toprule
    Pretrain seed & primary identity & single-level identity & support $\Delta_{\mathrm{joint}}$ & support $\Delta_{\mathrm{comp}}$ \\
    \midrule
    239 & 43.16 & 43.72 & $-5.20$ & $-1.35$ \\
    240 & 43.62 & 44.16 & $-0.16$ & +2.59 \\
    242 & 42.62 & 45.94 & $-2.09$ & +4.09 \\
    \midrule
    Mean & 43.13 & 44.61 & $-2.48$ & +1.78 \\
    \bottomrule
  \end{tabular*}
\end{table}

Table~\ref{tab:checkpoint} separates a replicated capability from a bounded structural result. The
external identity readout is reproduced by two JEPA loss-weighting recipes. By contrast, the support
residual beyond predicted identity and geometry changes sign across checkpoints and remains
unresolved in the pooled interval. This pattern motivates the compositional interpretation rather
than a relation-specific claim.

\paragraph{Artifact provenance.}
Authoritative aggregate files are listed below; the Sr3D endpoint also includes all three
100,000-replicate bootstrap arrays.
\begin{sloppypar}\footnotesize
\noindent\emph{Identity:}
reports/\allowbreak 2026-07-14\_arkit\_full\_endpoint/\allowbreak aggregate\_score.json\\
\emph{Support:}
reports/\allowbreak 2026-07-17\_sr3d\_relation\_rerun/\allowbreak aggregate\_score.json\\
\emph{Joint controls:}
reports/\allowbreak 2026-07-23\_sr3d\_joint\_control/\allowbreak aggregate\_joint\_control.json\\
\emph{Spatial reach:}
reports/\allowbreak 2026-07-24\_arkit\_scene\_global\_context/\allowbreak aggregate\_score.json
\end{sloppypar}
Endpoint manifests bind scene membership, query bytes, model/checkpoint identities, head states,
donor assignments, and score hashes. Raw semantic labels are unavailable to feature extraction and
head fitting for held-out partitions and are opened only by the final scorer.

\end{document}